\documentclass[]{interact}

\usepackage{epstopdf}
\usepackage{color}
\usepackage[numbers,sort&compress]{natbib}
\bibpunct[, ]{[}{]}{,}{n}{,}{,}
\renewcommand\bibfont{\fontsize{10}{12}\selectfont}
\makeatletter
\def\NAT@def@citea{\def\@citea{\NAT@separator}}
\makeatother

\theoremstyle{plain}

\theoremstyle{definition}

\theoremstyle{remark}

\begin{document}


\title{Bipedal Robot Running: Human-like Actuation Timing Using\\ Fast and Slow Adaptations}

\author{
\name{Yusuke Sakurai\textsuperscript{a}\thanks{CONTACT Tomoya Kamimura. Email: kamimura.tomoya@nitech.ac.jp} , Tomoya Kamimura\textsuperscript{a}, Yuki Sakamoto\textsuperscript{a}, Shohei Nishii\textsuperscript{a}, Kodai Sato\textsuperscript{a}, Yuta Fujiwara\textsuperscript{a}, and Akihito Sano\textsuperscript{a}}
\affil{\textsuperscript{a}Department of Electrical and Mechanical Engineering, Nagoya Institute of Technology, Aichi, Japan}
}

\maketitle

\begin{abstract}
We have been developing human-sized biped robots based on passive dynamic mechanisms. In human locomotion, the muscles activate at the same rate relative to the gait cycle during running. To achieve adaptive running for robots, such characteristics should be reproduced to yield the desired effect,  In this study, we designed a central pattern generator (CPG) involving fast and slow adaptation to achieve human-like running using a simple spring-mass model and our developed bipedal robot, which is equipped with actuators that imitate the human musculoskeletal system. Our results demonstrate that the CPG-based controller with fast and slow adaptations, and a adjustable actuator control timing can reproduce human-like running. The results suggest that the CPG contributes to the adjustment of the muscle activation timing in human running.
\end{abstract}


\section{Introduction}
We have been developing human-sized bipedal robots based on passive dynamic mechanisms, which plays a significant role in human dynamical locomotion.
McGeer~\cite{McGeer1990} developed a simple bipedal robot with passive legs attached to the hip; this robot could walk stably without the need for any energy input other than gravity by descending a slope.
The aforementioned study indicated that passive locomotion can play a significant role in gait.
However, the dynamical mechanisms under such locomotion is complex to fully understand sorely from observation. To overcome the limitations of observational approach, several researchers have investigated gait mechanisms using simple walking models~\cite{Garcia1998, Goswami1998, Collins2005, Ikemata2008, Kino2019, Okamoto2020, Kamimura2023}.
Bipedal locomotion consists of not only walking but also running. Therefore, some researchers further developed simple running models. For example, the spring-loaded inverted pendulum (SLIP) model, comprising a point mass and prismatic massless spring, can effectively reproduce the dynamics of running~\cite{Blickhan1989, Full1999, Geyer2005, Geyer2006, Clark2014, Gan2018}.
So far, to realize the running motion, we developed a bipedal robot that utilizes bouncing rod dynamics~\cite{Miyamoto2010}.
Recently, we developed a bipedal robot with actuators that could imitate the human musculoskeletal system, while utilizing the passive nature of the dynamics of the robot~\cite{Kamimura2021}.

Walking and running locomotion is executed through dynamic interactions between the nervous system, body, and environment.
Accordingly, bipedal robot controllers have been proposed based on human neurophysiology.
Reportedly, the centers involved in the execution of patterned locomotion in humans, such as walking and running, exist at the lower levels of the central nervous system (e.g., the spinal cord and brainstem), which are simulated via central pattern generators (CPGs)~\cite{Rybak2006, Cappellini2006, Ivanenko2006, Ivanenko2007, Molkov2015,Ichimura2023}.
CPGs employ a hierarchical network of rhythm generators that generate the rhythm of walking and running and pattern formulators, which are responsible for the timing of muscle activation.
Rybak et al.~\cite{Rybak2006} developed a model of mammalian spinal cord circuits and investigated the firing patterns of muscle activity. They incorporated CPGs obtained by observing muscle activity in cats whose spinal cord was disconnected from the higher centers, and the model can temporarily reset muscle activity upon sensory input from limbs.

Several researchers have investigated the emergence of gait through the interaction of CPGs, which are assigned to legs using dynamical models and legged robots~\cite{Taga1991,Manoonpong2009, Fukuoka2013, Aoi2017, Fujiki2018, Wang2020, Russo2023}.
Aoi and Tsuchiya~\cite{Aoi2006, Aoi2007} reported that stable walking can be achieved in a bipedal compass model by adaptively changing the gait period using a CPG that involves phase resetting, which is inspired by the rhythm resetting of human walking.
Recently, some researchers combined machine learning and CPG controller to reveal the optimal relationship between the CPG phase and motor control~\cite{Matsubara2006, Kasaei2021, Bellegarda2022, Herneth2023}.

Observably, humans fire their muscles at the same rate relative to the gait cycle during running~\cite{Perry2010}.
Thus, to achieve adaptive running using robots, such a characteristic should be reproduced.
When a CPG is used solely with phase resetting for running control, although the overall period can be adaptively varied, adjusting the timing of muscle firing is unachievable as yet.
Moreover, Fujiki et al.~\cite{Fujiki2015} reported that applying slow adaptation (late adaptation) to CPG based on the neurophysiological findings \cite{Morton2006} in addition to phase resetting for controlling a compact bipedal robot walking on a split-belt, with different velocities of the left and right limbs, the motion converges to a steady-state gait.
This study proposes an adaptive control system, which is based on the current knowledge of human neurophysiology, to achieve a stable running condition in a human-sized bipedal robot.
Unlike machine learning studies that explore optimized controllers for the environment, in this study, we designed the controller based on CPG aiming to reveal how the robot and controller manage to the environment by adaptive behavior.
We aimed to achieve adaptive running of the bipedal robot by adjusting the actuation timings via fast and slow adaptations.
We designed a CPG-based controller to manifest adaptive behavior by providing fast and slow adaptations.
Specifically, in the rhythm generator, the phase angular velocity is adjusted via fast adaptation (phase resetting) and slow adaptation to the actual gait cycle.
Furthermore, the pattern formulator adjusts the muscle activation timing on the basis of the leg-state feedback to yield a running pattern.
We validated the proposed control system through numerical simulations of a simple running model and experiments on a bipedal robot.

\section{Method}
In this study, we propose a control system based on a CPG to achieve adaptive actuation timings using fast and slow adaptation mechanisms.
We designed a rhythm generator that adapted to the gait cycle, which was determined by the interaction between the robot body and environment, and a pattern formulator that controlled the actuators based on the rhythm generator.

\subsection{Bipedal robot based on passive dynamics}
\begin{figure}[tb]
  \centering
  \includegraphics[width=120mm,angle=0]{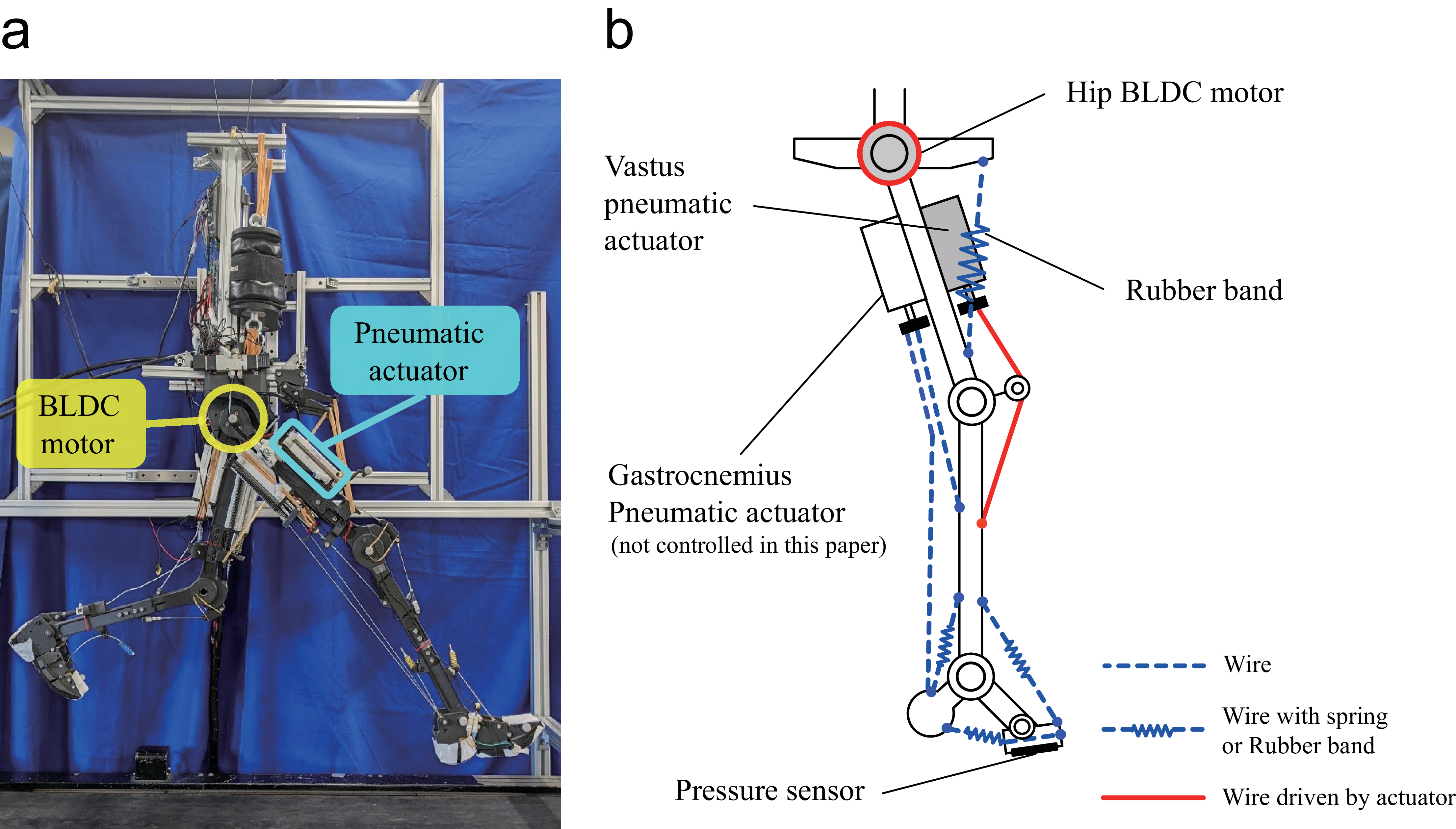}
  \caption{(a) Bipedal running robot. BLDC motors in hip joints actuate thigh links. Muscle-tendon system represented by pneumatic actuators actuates knee and foot links. (b) Schematics of robot. Links are connected with active (solid red lines) or passive (dashed blue lines) wires and springs. (Colored version is available online.)}
  \label{fig:bipedal_robot}
\end{figure}

The human-sized bipedal robot, developed in our laboratory, is illustrated in Fig.~\ref{fig:bipedal_robot} (leg length: 0.76~[m]; body mass: 20~[kg]).
The robot is composed of passive elements such as springs and rubber bands.
However, a minimum number of actuators are implemented because the energy input is a prerequisite for running on flat ground.
In particular, the vastus muscles, which are involved in knee extension, are represented by a wire connected to a pneumatic actuator (CDQ2A63-75DZ, AirTAC).
In addition, a brushless direct current (BLDC) motor (RMD-X8 Pro, MyActuator) is used to actuate the hip-joint extension.
To assist in thigh extension, a rubber band is connected between the pelvis and thigh to act as the thigh muscle. The touchdown timings are sensed using a pressure sensor (FSR406, Interlink Electronics) on the sole.
The robot is constrained in the sagittal plane.

\subsection{Rhythm generator with fast and slow adaptations}
\begin{figure}
	\centering
	\includegraphics[width=100mm]{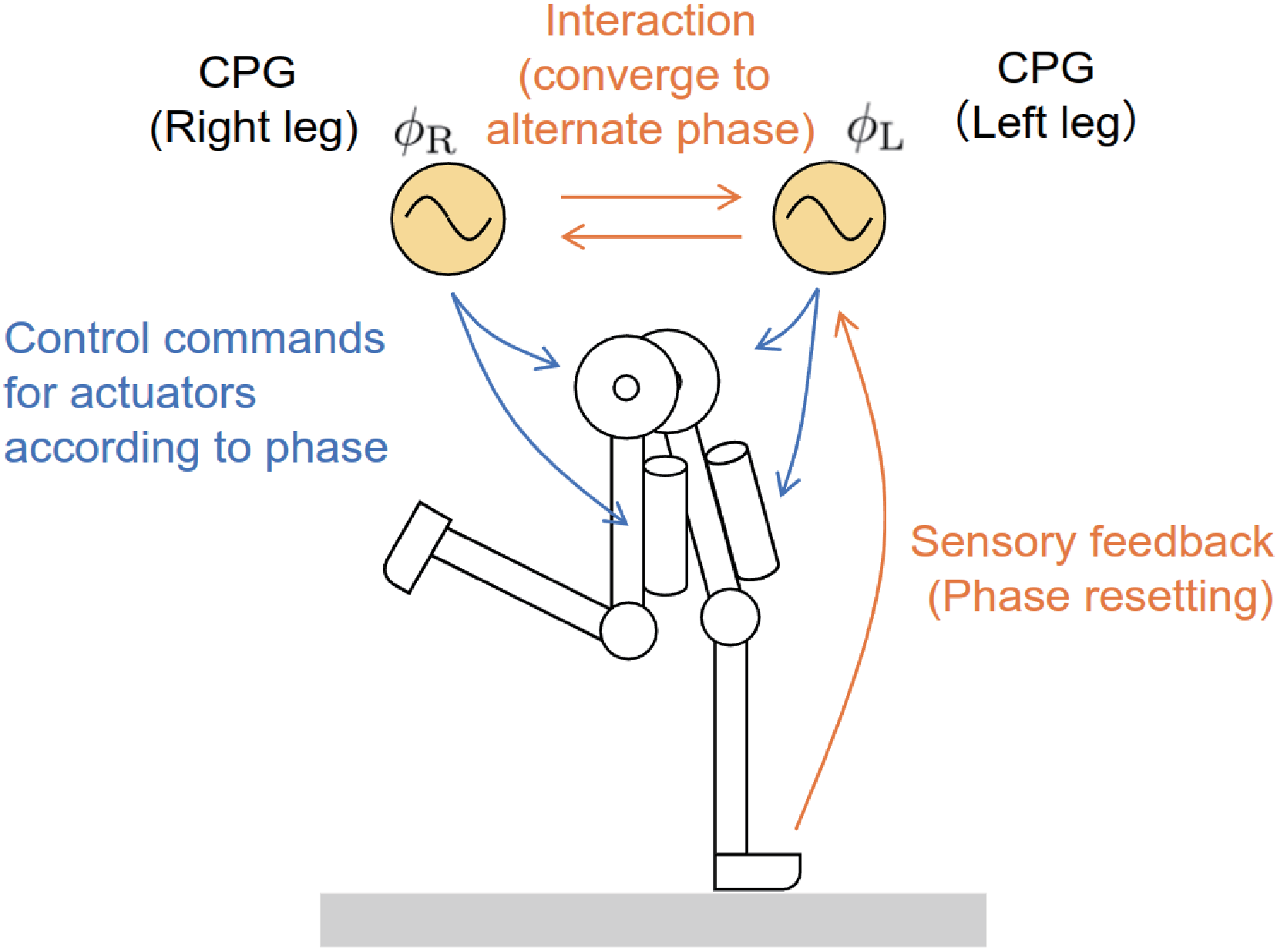}
	\caption{Schematics of controller. Left and right neural oscillators (CPGs) controls actuators according to their phases. Phases of CPGs are adjusted by sensory feedback and inter-limb interaction.}
	\label{fig:CPG_RG}
\end{figure}

We designed a rhythm generator involving fast and slow adaptations to adjust its cycle period to the actual gait cycle.
Moreover, this strategy guaranteed that the left and right legs move in opposite phases.
The rhythm generator was constructed as coupled oscillators of the left and right legs, where the phase oscillators ${\phi}_{\rm L}$ and ${\phi}_{\rm R}$ were defined for each leg as shown in Fig.~\ref{fig:CPG_RG}.

We defined the $n$th half-period (the time from the landing of one leg till the landing of the other leg) as $T_n$, and the half-period estimated by the phase oscillator as $T_{n}^{\rm e}$.
The basic angular velocity of the phase oscillator was defined as $\pi/T_n^{\rm e}$.
The period $T_{n}^{\rm e}$ of the robot's CPG does not always match the period $T_n$ of the actual running locomotion achieved by the robot.
As explained later, we designed the rhythm generator with slow adaptation to allow the estimated period $T_n^{\rm e}$ to converge to the actual period $T_n$, which was determined by the dynamical interaction between the body and the environment.
Because the left and right legs are designed to move in alternate phases, the target value of the left-right phase difference is $\pi$~[rad].
Furthermore, to adapt the oscillator period to the gait cycle immediately, we introduced phase resetting~\cite{Rybak2006} at the moment of touchdown.
Based on the design of the phase oscillator proposed by Aoi et al. \cite{Aoi2007}, the dynamics of each phase oscillator were determined as follows:

\begin{subequations}
\label{eq:phases_leftandright}
\begin{align}
    \dot{\phi}_{\rm R} &= \dfrac{\pi}{T_{n}^{\rm e}} + \varepsilon\sin\{(\phi_{\rm L}-\phi_{\rm R})-\pi\} - \phi^{\rm td}_{\rm R}\delta(t-t^{\rm td}_{\rm R}),\\
    \dot{\phi}_{\rm L} &=\dfrac{\pi}{T_{n}^{\rm e}} + \varepsilon\sin\{(\phi_{\rm R}-\phi_{\rm L})-\pi\}-\phi^{\rm td}_{\rm L}\delta(t-t^{\rm td}_{\rm L}),
\end{align}
\end{subequations}
where $\varepsilon$ denotes the gain of the phase difference, $\phi_{\rm L}^{\rm td}$ and $\phi_{\rm R}^{\rm td}$ represent the phases of the left and right oscillators at the touchdown moment, respectively, and $t_{\rm L}^{\rm td}$ and $t_{\rm R}^{\rm td}$ symbolize the moments when the touchdown occurs.
$\delta(\cdot)$ is Dirac's delta function.
The third term of each oscillator indicates the fast adaptation (phase reset) at the touchdown moment.

The estimated half-period $T_{n}^{\rm e}$ is designed to converge to the actual half-period $T_n$ by the following update rule as a slow adaptation by using the gain $K_{\rm p}$, $K_{\rm d}$, and the periodic difference $\Delta T_n=T_{n}^{\rm e}-T_{n}$ as

\begin{align}
    \label{eq:diffeq_est_halfPeriod}
    T_{n+1}^{\rm e}=T_{n}^{\rm e}-K_{\rm p} \Delta T_{n} -K_{\rm d}\frac{\Delta T_{n}-\Delta T_{n-1}}{T_{n}}.
\end{align}

Further, we defined the phase difference $\phi=\phi_{\rm R}-\phi_{\rm L}$ and phase sum $\psi=\phi_{\rm R}+\phi_{\rm L}$.
We assumed that running was initiated upon touchdown of the right leg in our experiment.
Therefore, an odd number of $n$ indicates the touchdown of the right leg, and an even number of $n$ indicates the touchdown of the left leg.

\subsection{Pattern formulator for reproducing human-like actuation timing}
To achieve human-like running, we designed a pattern formulator, which was based on the muscle activity during human running~\cite{Perry2010}.
Based on the phase angles of the rhythm generator, the pattern formulator switches the control laws of the hip and vastus pneumatic actuators.

\begin{figure}
    \centering
    \includegraphics[width=\linewidth]{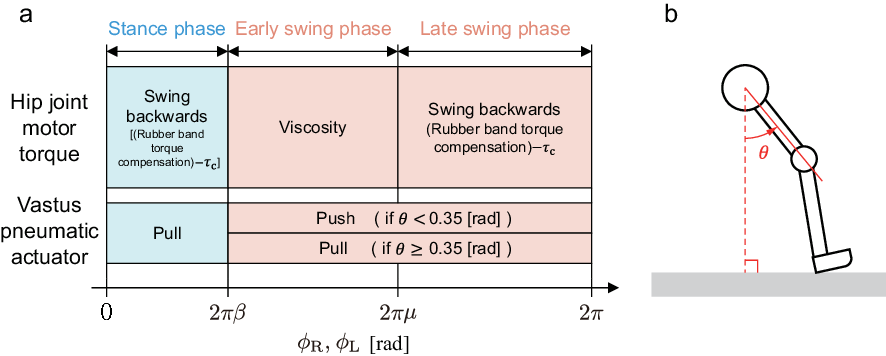}
    \caption{(a) Phases of rhythm generator are categorized into three phases: stance, early swing, and late swing. Pattern formulator switches control laws of actuators (hip joint motor and vastus pneumatic actuator) according to phase angle $\phi_i$ ($i = \mathrm{L, R}$). (b) Thigh angle $\theta$ is defined as angle between thigh link and vertical direction.}
    \label{fig:CPG_PF}
\end{figure}

The phases are categorized into three ranges, as illustrated in Fig.~\ref{fig:CPG_PF}a.
The stance phase is defined as the phase range $0 \leq \phi_i < 2\pi\beta$~[rad] ($i={\rm L, R}$), and the swing phase is defined as the phase range $2\pi\beta \leq \phi_i < 2\pi$~[rad], where $\beta$ indicates the duty rate, which is determined using measured human data~\cite{Perry2010}.
Furthermore, we categorized the swing phase into two phases: the early swing phase ($2\pi\beta \leq \phi_i < 2\pi\mu_N$~[rad]) and the late swing phase ($2\pi\mu_N \leq \phi_i <2\pi$~[rad]), where the rate $\mu_N$ is defined hereafter.

The output torque of the hip motor is switched according to the phase.
In the stance phase, the leg exerts a force (i.e. kicks) on the ground because the motor outputs a constant torque $\tau_{\rm c}$ while compensating for the torque generated by the rubber bands.
During the early swing phase, a rubber band is used to swing the leg forward.
In the late swing phase, the motor outputs a torque $\tau_{\rm c}$ to retract the legs.

The pneumatic actuator switches the output according to the phase and state of the leg.
The thigh angle $\theta$ is defined as the current angle of the thigh link with respect to the vertical downward direction of the robot, where the swing-forward direction is positive (Fig.~\ref{fig:CPG_PF}b).
For the simplicity of the control, $\theta$ is utilized as a leg angle instead of the leg axis angle.
During the stance phase, the pneumatic actuator tightens the vastus muscle wire.
When the stance phase of the rhythm generator terminates or the leg is lifted off the ground, the pneumatic actuator relaxes the vastus muscle wires to prevent the robot from hitting the ground.
When the thigh angle $\theta$ exceeded a specific angle (we used 0.35~[rad] ($=20$~[deg]) in the experiments), the vastus muscle wire was slightly tightened in preparation for the subsequent landing.

To maintain a steady running movement, a constant thigh-swing angle must be maintained regardless of the gait cycle.
To achieve this condition, we introduced an actuation timing adjustment mechanism into the pattern formulator in addition to the adaptation in the rhythm generator.
Physiological studies have shown that sensory information is not only input to the rhythm generator, but also to the pattern formulator~\cite{Rybak2006}.
In the current study, the pattern formulator was designed to adjust actuation timing each period in order to achieve the desired thigh angle at ground contact.
Specifically, adjusting the rate $\mu_N$ appropriately using the feedback of the thigh swing angle is a prerequisite for properly adjusting the time between touchdown and the beginning of the late swing phase.
To follow the target angle $\theta_{\rm d}^{\rm s}$, we measured the thigh angle $\theta^{\rm s}_N$ at the beginning of the late swing phase and calculated $\mu_{N+1}$ by using the gain $K_{\mu}$ as follows:
\begin{align}
    \label{eq:mu_variable}
	\mu_{N+1}=\mu_{N}-K_{\mu}(\theta_{N}^{\rm s}-\theta_{\rm d}^{\rm s}) .
\end{align}
Furthermore, in accordance with the relationship between the actual swing angle $\theta_{N}^{\rm s}$ and target swing angle $\theta_{\rm d}^{\rm s}$, the value of $K_\mu$ is switched as follows:
\begin{align}
    \label{eq:K_mu}
    K_\mu = \begin{cases}
        0.005, &(\theta_{N}^{\rm s}>\theta_{\rm d}^{\rm s})\\
        0.1. &(\theta_{N}^{\rm s}\leq\theta_{\rm d}^{\rm s})
    \end{cases}
\end{align}
This condition yields a strong adjustment when the thigh-swing angle $\theta_N$ is smaller than the target value $\theta_{\rm d}$.
In the experiment, we set $\theta^{\rm s}_{\rm d}=0.61$~[rad] ($=35$~[deg]) based on the human running and $K_{\rm p}=0.1$.
Note that in human running~\cite{Perry2010}, the hip joint switches from flexion to extension when the leg angle to the vertical direction is about 0.56~[rad] ($=32$~[deg]).

\subsection{Simple model for verification of rhythm generator adaptation}

\begin{figure}[tb]
    \centering
	\includegraphics[scale = 1.0]{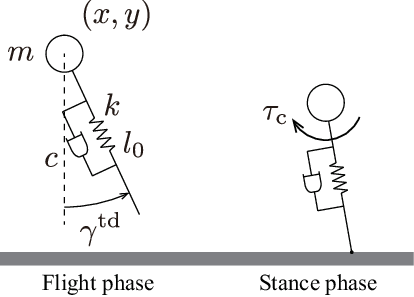}
	\caption{Simple SLIP model with leg dumper and hip actuator.}
	\label{fig:SLIP}
\end{figure}

Before real robot experiments, we conducted numerical simulations to confirm the validity of our controller, especially focusing on the slow adaptation of the estimated half-period expressed by eq.~\eqref{eq:diffeq_est_halfPeriod} using a simple running model.
We used a spring-loaded inverted pendulum (SLIP) model, which consists of a point mass $m$ and a massless leg (Fig.~\ref{fig:SLIP}). The leg is represented by a prismatic spring $k$ and dumper $c$. The horizontal and vertical positions of the mass are represented by $x$ and $y$, respectively. The gravitational acceleration is $g$. When the leg is in the air, its length remains $l_0$ and its angle keeps the touchdown angle $\gamma^{\rm td}$. When the tip of the leg reaches the ground, it is constrained on the ground and behaves as a frictionless joint. When the stance leg returns to its nominal length after compression, the tip leaves the ground, and the leg angle immediately returns to the touchdown angle. In the stance phase, the leg is actuated by the hip motor constant torque $\tau_{\rm c}$.

The equation of motion is given by
\begin{align}
    \label{eq:simple_model}
    M\ddot{\bm{q}}+D(\bm{q}, \bm{\dot{q}})
    +G(\bm{q})=\bm{T},
\end{align}
where $\bm{q} = [x,y]^\top$. $M$, $D(\bm{q},\bm{\dot{q}})$, $G(\bm{q})$, and $\bm{T}$ represent inertia matrix, dumping terms, conservative forces, and actuator input, respectively. (For details, see Appendix.~A.)
According to human parameter~\cite{Geyer2006}, we set $m=50$~[kg], $l_0=1$~[m], $k=8000$~[N/m], $c=20$~[Ns/m], and $g=9.8$~[m/s$^2$].
Furthermore, we heuristically set $\varepsilon=4$, $\beta=0.6$, $\tau_{\rm c}=52$~[Nm], and $\gamma^{\rm td}=\pi/6$~[rad].
Because our model involves a massless leg, control torque terms related to leg inertia (Fig.~\ref{fig:CPG_PF}) are omitted in the simulation.
When the step number is odd, the right leg CPG controls the leg, and when it is even, the left CPG controls the leg.

\section{Results}
\subsection{Effect of rhythm generator adaptation on simple model}
\begin{figure}[tb]
    \centering
	\includegraphics[width = \linewidth]{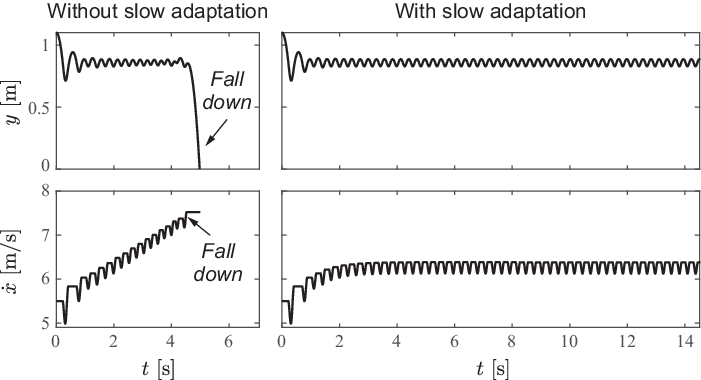}
        \caption{Time profiles of $y$ and $\dot{x}$ of SLIP model. Left and right figures indicate results for $(K_{\rm p},K_{\rm d})=(0,0)$ (without slow adaptation) and $(0.8,0.1)$ (with slow adaptation), respectively. Model falls down around 5~[s] without slow adaptation.}
        \label{fig:x_y_simulation}
\end{figure}

First of all, we verified the validity of fast and slow adaptation of the rhythm generator using numerical simulation of the simple model.
We conducted simulations changing slow-adaptation gains $K_{\rm p}$ and $K_{\rm d}$ defined in eq.~\eqref{eq:diffeq_est_halfPeriod}.
The simulation results for $(K_{\rm p},K_{\rm d})=(0,0)$ (without slow adaptation) and $(K_{\rm p},K_{\rm d})=(0.8,0.1)$ (with slow adaptation) are shown in Fig.~\ref{fig:x_y_simulation}.
Although the model fell down in 17 steps without slow adaptation, it attracted to a certain stable limit cycle which resulted into running over 3000 steps with slow adaptation.

\begin{figure}[tb]
    \centering
	\includegraphics[scale = 1.0]{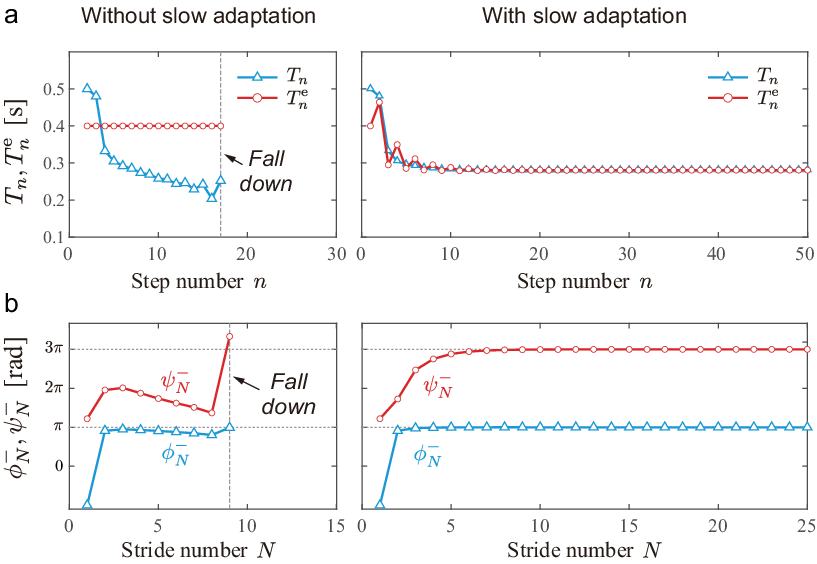}
	\caption{Responses of CPG for $(K_{\rm p},K_{\rm d})=(0,0)$ (without slow adaptation) and $(0.8,0.1)$ (with slow adaptation), respectively. (a) Responses of $T_n$ (triangle) and $T_n^{\rm e}$ (circle) to step number $n$. (b) Responses of phase difference $\phi_N^-$ (triangle) and phase sum $\psi_N^-$ (circle) to stride number $N=(n+1)/2$ and $*_{N}^{-}$ indicates the phases immediately before the $N$th touchdown of the right leg. We used the stride number $N$ instead of the step number $n$ to focus solely on the touchdown moments of the right leg.}
    \label{fig:sim_CPG}
\end{figure}

The simulation results of step duration are shown in Fig.~\ref{fig:sim_CPG}a, where the initial estimated value is set as $T_0^{\rm e}=0.4$~[s].
Although the estimated half-period $T_n^{\rm e}$ was not adjusted in the case without slow adaptation, $T_n^{\rm e}$ and actual half-period $T_n$ asymptotically approached and converged to each other in the case with slow adaptation.

Fig.~\ref{fig:sim_CPG}b presents the response of phase difference $\phi_N^-$ and phase sum $\psi_N^-$ to stride number $N$, where $N=(n+1)/2$ and $*_{N}^{-}$ indicates the phases immediately before the $N$th touchdown of the right leg.
Herein, we used the stride number $N$ instead of the step number $n$ to focus solely on the touchdown moments of the right leg.
Notably, the stride refers to the footfall number focusing only on the right leg.
Although $\phi_N^-$ and $\psi_N^-$ did not converge in the case without slow adaptation, they converged to $\pi$~[rad] and $3\pi$~[rad], respectively, in the case with slow adaptation.

Therefore, by adjusting $T_n^{\rm e}$ using slow adaptation, the phase difference approaches the opposite phase $\pi$~[rad] and the phase sum approaches $3\pi$~[rad] as $N$ increases, indicating that the phase has been adjusted to increase by exactly $2\pi$~[rad] during one cycle.

The aforementioned results demonstrate that the estimated half-period $T_n^{\rm e}$ can be adapted to the actual period via slow adaptation while guaranteeing the left-right opposite phase.
Therefore, even if there remains a mismatch between the actual period and the initial value of the estimated half-period, muscle activation can be effectively performed at the same timing like humans using the slow adaptation.

\subsection{Robot experiments}
\subsubsection{Fast and slow adaptation of rhythm generator}
To verify the effect of fast and slow adaptation of the estimated half-period, we performed experiments on the robot for slow adaptation gain $K_{\rm p}=0$ and $0.1$ in eq.~\eqref{eq:diffeq_est_halfPeriod}, respectively.
For simplicity, we set $K_{\rm d}=0$ in the experiments.
In these experiments, we assumed the absence of a feedback of the thigh angle in the pattern formulator, which was achieved by setting $K_\mu = 0$.

For the rhythm generator, we set $\varepsilon = 4$ and $T_0^{\rm e} = 550$~[ms].
The value of $\varepsilon$ was set based on some preliminary experiments. When $\varepsilon$ is set too small, the convergence of the relative phase $\phi$ is too slow, causing the robot's left and right legs to be unable to move alternately, ultimately leading to a fall.

For the pattern formulator, based on human parameter and several preliminary experiments, we set $\beta = 0.25$, $\mu=0.6$, and $\tau_{\rm c} = 3$~[Nm].
Note that the duty rate in human running is about 35~\% and the hip extensor muscle groups begin to contract after 50~\% of the gait cycle~\cite{Perry2010}.
The treadmill speed during the experiment was set at 10~[km/h].

\begin{figure}[tb]
	\centering
	\includegraphics[scale = 1.0]{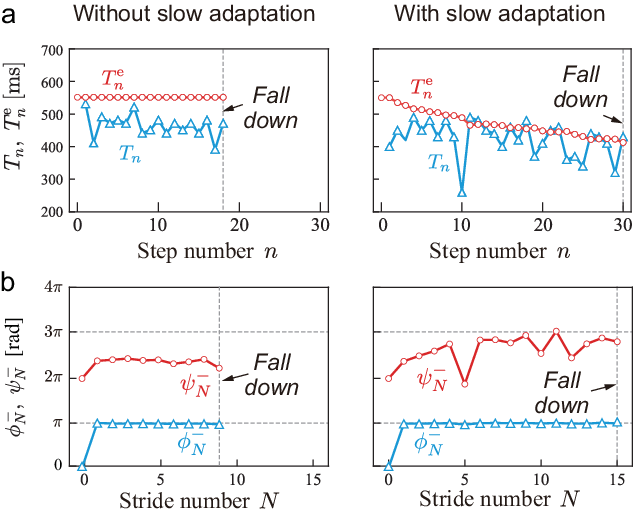}
	\caption{Responses of rhythm generator. (a) Actual and estimated half-period $T_n$ (triangle) and $T_n^{\rm e}$ (circle) to step number $n$. (b) Phase difference $\phi_N^-$ (triangle) and phase sum $\psi_N^-$ (circle) at touchdown moments of right leg to stride number $N = (n+1)/2$. Left and right figures indicate $K_{\rm p}=0$ (without slow adaptation) and $K_{\rm p}=0.1$ (with slow adaptation), respectively.}
	\label{fig:exp1_CPG}
\end{figure}

Fig.~\ref{fig:exp1_CPG}a presents a plot of the response of the estimated half-period $T_n^{\rm e}$ to the number of steps $n$.
For $K_{\rm p}=0$, $T_n^{\rm e}$ maintained its initial value of 550~[ms], regardless of the actual half-period $T_n$ because the estimated half-period was not adjusted.
In contrast, for $K_{\rm p}=0.1$, the estimated half-period $T_n^{\rm e}$ and actual half-period $T_n$ asymptotically converge to each other, same as the simple model simulation (Fig.~\ref{fig:sim_CPG}a).
Note that, in both experiments, the fast adaptation enabled running, even when there remained a mismatch between $T_n^{\rm e}$ and $T_n$.

Fig.~\ref{fig:exp1_CPG}b depicts the change of phase difference $\phi_N^-$ and phase sum $\psi_N^-$ at the moment immediately prior to the touchdown of the right leg with respect to $N$.
For both $K_{\rm p}=0$ and $0.1$, the phase difference $\phi_N^-$ converges to $\pi$~[rad].
On the other hand, while $\psi_N^-$ remained at approximately $2.5\pi$~[rad] for $K_{\rm p}=0$, it increased to approximately $3\pi$~[rad] for $K_{\rm p}=0.1$.

\begin{figure}[tb]
	\centering
	\includegraphics[scale=1.0]{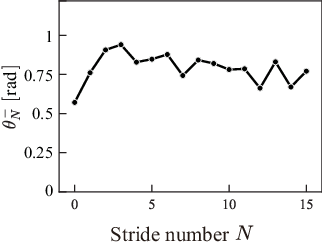}
	\caption{Response of swing angle of thigh $\theta^-_N$ at touchdown moments of right leg to stride number $N$. $\theta^-_N$ gradually decreased as $N$ increased.}
	\label{fig:exp1_thighangle}
\end{figure}

\begin{figure}[tb]
	\centering
	\includegraphics[width=140mm]{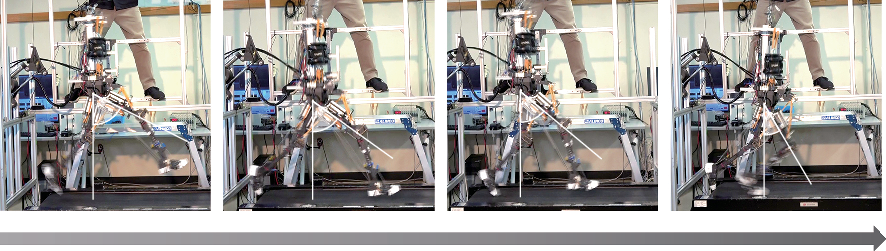}
        \caption{Snapshots of timings immediately before right leg touch down. The touchdown angle gradually reduced and resulted in falling down.}
	\label{fig:mu_const_snap}
\end{figure}

Although the robot achieved sustained running until $N=15$, it could not achieve steady running and fell after running.
The response of thigh angle $\theta_N^-$ to $N$ is plotted in Fig.~\ref{fig:exp1_thighangle}.
Fig.~\ref{fig:mu_const_snap} shows the snapshots of the experiment for $13 \leq N \leq 16$.
Even when the slow adaptation in the rhythm generator is activated, the robot cannot continue to run because the swing angle of the thigh gradually decreases as $N$ increases, which resulted in a stumble.

\subsubsection{Pattern formulator adjustment}

To elucidate the effect of the actuation timing adjustment in the pattern formulator, we conducted robot experiments with adjustment of the swing rate $\mu_N$, which is determined using equations~\eqref{eq:mu_variable} and~\eqref{eq:K_mu}.
From this experiment, we verified that a longer period of sustained running can be achieved by adjusting $\mu_N$ on the basis of the thigh-angle feedback in the pattern formulator.

\begin{figure}[tb]
	\centering
	\includegraphics[scale = 1.0]{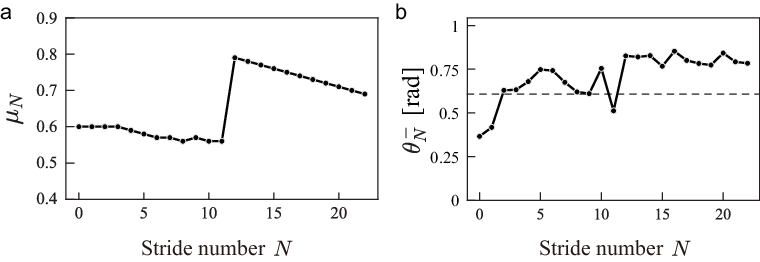}
	\caption{Response of (a) $\mu_N$ and (b) thigh swing angle $\theta^-_N$ to stride number $N$ with $\mu_N$ adjustment. Dashed line indicate control desired thigh angle $\theta^{\rm s}_{\rm d} = 0.61$~[rad] in eq.~\eqref{eq:K_mu}.}
	\label{fig:exp2_PF}
\end{figure}

Fig.~\ref{fig:exp2_PF}a presents the response of $\mu_N$ to the stride number $N$.
$\mu_N$ decreases and converges to approximately 0.55 around $N=6$ to $12$.
When $N=12$, $\mu_N$ increases substantially to an approximate value of 0.80 and subsequently decreases again.

\begin{figure}[tb]
	\centering
	\includegraphics[width=140mm]{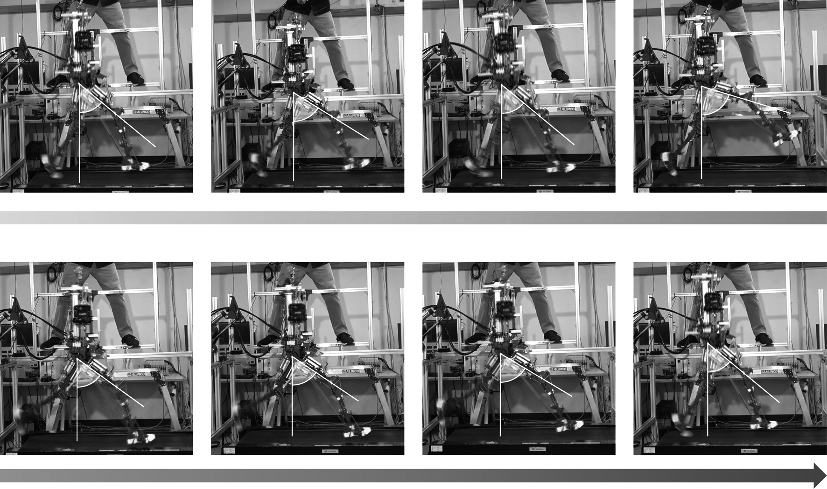}
        \caption{Snapshots of timings immediately before right leg touch down with adaptive $\mu_N$. The touchdown angles are kept almost constant.}
	\label{fig:mu_var_snap}
\end{figure}

Fig.~\ref{fig:exp2_PF}b shows the plot of the right-thigh angle $\theta_N^-$ with respect to $N$.
The swing angle decreases regardless of the value of $\mu_N$ until $N=9$.
In contrast, after $N=10$, although the swing angle continues to decrease when $\mu_N$ is constant, it increases with $N$ when $\mu_N$ is adaptive.
Moreover, when the swing angle $\theta_N^-$ took values below $\theta^{\rm s}_{\rm d} = 0.61$~[rad] at $N=11$, it was affected by strong feedback shown in eq.~\eqref{eq:K_mu} and took apparently larger angle in the next stride.
Fig.~\ref{fig:mu_var_snap} shows the snapshots for eight strides. Thanks to the $\mu_N$ adjustment, the thigh angle is kept almost constant compared to the experiment with constant $\mu_N$ (Fig.~\ref{fig:mu_const_snap}), which contribute to keep running.
The obtained results reveal that the adaptive $\mu_N$ prevents the monotonic decrease in the thigh swing angle, thereby resulting in continuous running of more than $N=20$.

\begin{figure}[tb]
	\centering
	\includegraphics[scale = 1.0]{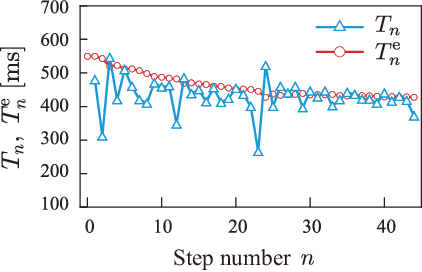}
	\caption{Response of $T_n$ (triangle) and $T_n^{\rm e}$ (circle) to $n$ with adjustable $\mu_N$.}
	\label{fig:exp2_CPG}
\end{figure}

The response of $T_n^{\rm e}$ to $n$ is plotted in Fig.~\ref{fig:exp2_CPG}.
When $\mu_N$ is adjustable, the decrement rate of $T_n^{\rm e}$ is suppressed, and $T_n^{\rm e}$ converges to a constant value ($\approx 440$~[ms]) as $n$ increases.

\begin{figure}[tb]
	\centering
	\includegraphics[scale = 1.0]{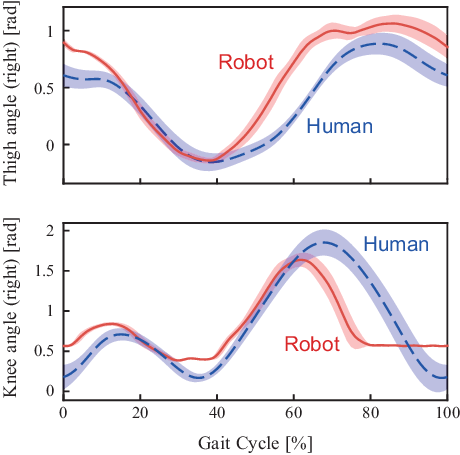}
	\caption{Comparison of time profiles of leg link angles (thigh (top) and knee (bottom)) between robot and human experiment. Red solid and blue dashed lines indicate averaged robot (5 strides) and human results (10 strides), respectively. Red and blue area indicates standard deviation.}
	\label{fig:exp2_leg_angle}
\end{figure}

To verify the validity of the proposed controller, we compared kinematics of the robot experiment and human running at the similar velocity.
Fig.~\ref{fig:exp2_leg_angle} shows time profiles of thigh and knee angles of the robot and human running.
While our robot achieved running with 10~[km/h] ($Fr = 1.02$), human running data~\cite{Fukuchi2017} was taken with 12.6~[km/h] ($Fr = 1.18$), where $Fr = v/\sqrt{gl}$ indicates Froude number, and $v$ and $l$ indicates forward velocity and leg length, respectively. We set human leg length $l=1.0$~[m].
Both robot and human joints showed similar behavior.
In particular, both joint angles take their peaks almost the same timings.
The joint angle values were almost similar, however, the maximum value of the thigh angle of the robot was 0.26~[rad] larger than that of human, and the minimum value of the knee angle was 0.35~[rad] larger than that of the human.
These are supposed to be due to the parameters of the passive parts of the robot differing from those of the human. Specifically, the rubber bands in the thighs were too strong to extend, and the elasticity of the gastrocnemius and other muscles was too strong to extend the knees fully.

\section{Discussion}
\subsection{Effect of estimated gait cycle adaptation in rhythm generator}
In this study, to realize a bipedal robot capable of running with human-like actuation timing, we designed a controller based on the CPG that could adapt its gait cycle through slow adaptation, in addition to fast adaptation (phase resetting) in the rhythm generator.
While the fast adaptation adjusts the phase impulsively at each foot contact, the slow adaptation adjusts the period with each step. From a time-scale perspective, the running period is about the same as for slow adaptation cycle, whereas fast adaptation occurs in a very short duration. This relationship is consistent with human physiological study~\cite{Morton2006}.
Notably, the results validated that both the simple model and the robot could achieve continuous running by adaptive controller, although their body dynamics, actuator output, and resulting gait cycle and duty rate were largely different.
Moreover, owing to the slow adaptation of the rhythm generator, the estimated gait cycle and the actual gait cycle asymptotically approached each other through slow adaptation (Figs.~\ref{fig:sim_CPG}a and~\ref{fig:exp1_CPG}), whereas the left and right legs were actuated in alternate phases by employing the phase-difference compensation term.
Furthermore, fast adaptation allowed the robot to run even for a mismatch between the estimated and actual gait cycles.
Particularly, in the simple model, a stable limit cycle was formed from the interaction between nervous system, body, and environment, and the locomotion was attracted to the stable limit cycle.
These results suggest that not only the fast adaptation but also slow adaptation of the rhythm generator plays an significant role to generate the adaptive locomotion.

In the real robot experiment, the pattern formulator actuated the pneumatic actuators, which represent the vastus muscles of the robot, and the BLDC motors, which represent the hip extensor muscle groups, according to the phase angle of the rhythm generator.
Therefore, the obtained results demonstrated that the robot could run with a constant rate of muscle firing relative to the gait cycle regardless of the actual running cycle, which is similar to human running, owing to the fast and slow adaptation of the rhythm generator.

However, when we solely utilized adaptation in the rhythm generator, the running motion could not be sustained for a long time.
The thigh-swing angle decreased as the number of steps increased, thereby leading to this inability, as illustrated in Figs.~\ref{fig:exp1_thighangle} and~\ref{fig:mu_const_snap}.
In particular, this result was presumably a result of the constant $\mu_N$ used in the pattern formulator.
When the initial value of the estimated half-period $T_n^{\rm e}$ is larger than the actual half-period $T_n$, $T_n^{\rm e}$ decreased as $n$ increased owing to slow adaptation, where the phase angular velocity of the rhythm generator increased.
This resulted in a short stance phase, and the robot could not kick the ground sufficiently.
Moreover, the swing angle also decreased because the duration of the early swing phase was reduced.
In particular, the repetition of such a sequence is expected to decrease the swing angle of the thighs.
This experimental result suggested that a further adaptation mechanism is required to achieve continuous running.

\subsection{Effect of actuation timing adjustment in pattern formulator}
We achieved the continuous human-like running of our bipedal robot for approximately $20$ strides by adjusting mechanism of the actuation timing $\mu_N$ in the pattern formulator in addition to the two time scale adaptations in the rhythm generator.
From a time-scale perspective, the actuation timing adjustment cycle is the same as the running period, similar to the slow adaptation in the rhythm generator.
The robot's fall was prevented by adjusting $\mu_N$ appropriately (Fig.~\ref{fig:exp2_PF}a) when the thigh swing angle temporarily decreased to an inappropriate value, as shown in Figs.~\ref{fig:exp2_PF}b and~\ref{fig:mu_var_snap}, thereby yielding the successful result.
Furthermore, $\mu_N$ converged to a certain value of approximately 0.55 for $N=6$ to $11$, as depicted in Fig.~\ref{fig:exp2_PF}a.
This result suggests the existence of an appropriate $\mu_N$ for running, generated by the dynamic interaction between the body of the robot and the environment.
Moreover, the resulting running locomotion was similar to human running not only qualitatively but also quantitatively (Fig.~\ref{fig:exp2_leg_angle}), although the absolute values of the angles were slightly different due to the parameter difference between robot and human body.
Particularly, the timings at which the joint angles took peak values were almost identical to those of the human data.
Therefore, the obtained results suggest that humans invariably activate their muscles at the same rate with respect to the gait cycle using the appropriate muscle activation timing, which is defined by the dynamic interaction between the human body and the environment.

However, even when we used the pattern formulator with actuation timing adjustment, the robot could not maintain continuous running for a long period.
We discovered that the thigh angle at liftoff timing gradually decreased during running, thereby causing difficulties for the robot to kick the ground.
This trend resulted in the robot's inability to lift itself upward sufficiently, and the robot consequently stumbled and fell onto the treadmill.
As indicated by the results in Fig.~\ref{fig:exp2_PF}b, $\theta_N^-$ was maintained at approximately 50~[deg] after $N=12$, following the considerable change in $\mu_N$.
To achieve longer periods of continuous running, the duty rate $\beta$ in the pattern formulator should be accordingly adjusted, which will result in an appropriate swing angle.

\section{Conclusion}
In this study, we designed a CPG-based controller involving fast and slow adaptation and a adjustable actuator control timing to achieve human-like running motion of a bipedal robot.

Consequently, the CPG adapted to the actual period and achieved continuous running for both the simple model and the actual bipedal robot utilizing the fast and slow adaptation mechanism.
Furthermore, the results suggested that the CPG contributed to adjusting the muscle firing timing properly according to the state of the body, which could also result in optimum timing for human running.

However, in the robot experiment, even upon the implementation of the CPG with fast and slow adaptation in the rhythm generator and the actuation timing adjustment in the pattern formulator, the swing angle of the thigh decreased as the number of steps increased, and the robot was unable to achieve continuous running for a long period.
In future research, we will adjust the duty ratio $\beta$ on the basis of the swing angle.

Moreover, we will aim to improve upon the developed running robot and develop a walking robot to elucidate its relationship to adaptive human locomotion.
Thus, further studies in this research direction will lead to engineering applications not only for improved bipedal robots but also for walking assistive devices.

\section*{Acknowledgement}
This work was supported in part by JSPS KAKENHI Grant Number JP21K14104 and JP22H01445.

\bibliographystyle{tfnlm}
\bibliography{main}

\appendix
\section{Equation of motion of the simple model}
The terms in the equation \eqref{eq:simple_model} is given as follows:
\begin{align}
    M&=\begin{bmatrix}
        m & 0\\
        0 & m
    \end{bmatrix},\\
    D(\bm{q},\dot{\bm{q}}) &=
    \begin{cases}
        \begin{bmatrix}
            0\\
            0
        \end{bmatrix}, & \rm{in\ flight\ phase}\\
        \begin{bmatrix}
        c\cfrac{\hat{x} \dot{x}+y\dot{y}}{\hat{x}^2+y^2} \hat{x}\\
        c\cfrac{\hat{x} \dot{x}+y\dot{y}}{\hat{x}^2+y^2} y
        \end{bmatrix}, & \rm{in\ stance\ phase}
    \end{cases}\\
    G(\bm{q}) &=
    \begin{cases}
        \begin{bmatrix}
            0\\
            mg
        \end{bmatrix}, & \rm{in\ flight\ phase}\\
        \begin{bmatrix}
        k\left(\sqrt{\hat{x}^2+y^2}-l_0\right) \cfrac{\hat{x}}{\sqrt{\hat{x}^2+y^2}}\\
        mg+k\left(\sqrt{\hat{x}^2+y^2}-l_0\right) \cfrac{y}{\sqrt{\hat{x}^2+y^2}}
        \end{bmatrix}, & \rm{in\ stance\ phase}
    \end{cases}\\
    \bm{T} &=
    \begin{cases}
        \begin{bmatrix}
            0\\
            0
        \end{bmatrix}, & \rm{in\ flight\ phase}\\
        \begin{bmatrix}
        \cfrac{y}{\hat{x}^2+y^2}\tau_\mathrm{c}\\
        -\cfrac{\hat{x}}{\hat{x}^2+y^2}\tau_\mathrm{c}
        \end{bmatrix}, & \rm{in\ stance\ phase}
    \end{cases}
\end{align}
where $\hat{x} = x-x_0$ and $x_0$ indicates the toe touchdown position in stance phases.

\end{document}